# Using In-Context Learning for Automatic Defect Labelling of Display Manufacturing Data

*Babar Hussain\*, Qiang Liu\*, Gang Chen\*\*, Bihai She\*\*, Dahai Yu\**

*TCL Corporate Research, Hong Kong, S.A.R. China

**TCL China Star Optoelectronics Technology Co., Ltd., Shenzhen, China

**Abstract**

*This paper presents an AI-assisted auto-labeling system for display panel defect detection that leverages in-context learning capabilities. We adopt and enhance the SegGPT architecture with several domain-specific training techniques and introduce a scribble-based annotation mechanism to streamline the labeling process. Our two-stage training approach, validated on industrial display panel datasets, demonstrates significant improvements over the baseline model, achieving an average IoU increase of 0.22 and a 14% improvement in recall across multiple product types, while maintaining approximately 60% auto-labeling coverage. Experimental results show that models trained on our auto-labeled data match the performance of those trained on human-labeled data, offering a practical solution for reducing manual annotation efforts in industrial inspection systems.*

## 1. Introduction

The development of robust and reliable automated inspection systems is critical in modern display panel manufacturing to maintain high yields and product quality. Traditionally, these systems have relied on skilled human inspectors for defect detection and classification, a process that is both time-consuming and prone to inconsistencies [1]. While image processing-based rule systems have offered partial automation, they often fall short in handling the complexities of real-world defect patterns, necessitating further verification and re-repair by human operators [2]. Recent advancements in deep learning have shown promise in automating these tasks, improving accuracy and efficiency while enabling manufacturers to gain valuable insights for continuous improvement [3, 4, 5].

A key challenge in developing AI-driven image inspection systems lies in the need for extensive, high-quality training data. The manual annotation of this data requires significant expertise and resources, making it both time-consuming and prone to human error. Consequently, deep learning models often operate with limited data, leading to sub-optimal performance. While techniques like generative models offer a means to synthetically augment training data [4], the complexity involved in designing these systems and their limited applicability across diverse product types often make them less desirable.

Recent advances in artificial intelligence have introduced in-context learning (ICL), a capability that allows AI models to perform new tasks through examples, opening new possibilities in industrial automation. Models such as Painter [6] leverage ICL to perform diverse image tasks like depth estimation, segmentation, and image enhancement by reformulating them as image inpainting task. SegGPT [7] builds on this, demonstrating ICL's power in various types of segmentation tasks by enabling a single model to segment arbitrary objects and scenes based on provided examples.

Building on these developments, we propose an AI-assisted auto-labeling methodology that leverages ICL to learn from human annotations to automatically label unannotated data, thereby enhancing both annotation efficiency and training data quality. To further optimize the ICL model's performance, we introduce two key enhancements: (1) a training strategy optimized for display panel data characteristics and (2) a prompt design specifically tailored for the auto-labeling task.

To improve model generalization, we implement a comprehensive data augmentation strategy that creates composite images with multiple defects at varying scales and positions, addressing the limitations of single-defect datasets. We enhance prompt faithfulness through selective label suppression during training, ensuring the model segments only specified defect classes. Additionally, we streamline the annotation process through a scribble-based prompting mechanism, allowing simple user inputs instead of complete masks. This is achieved by incorporating a specialized token in the model architecture and training with paired scribble-to-mask data.

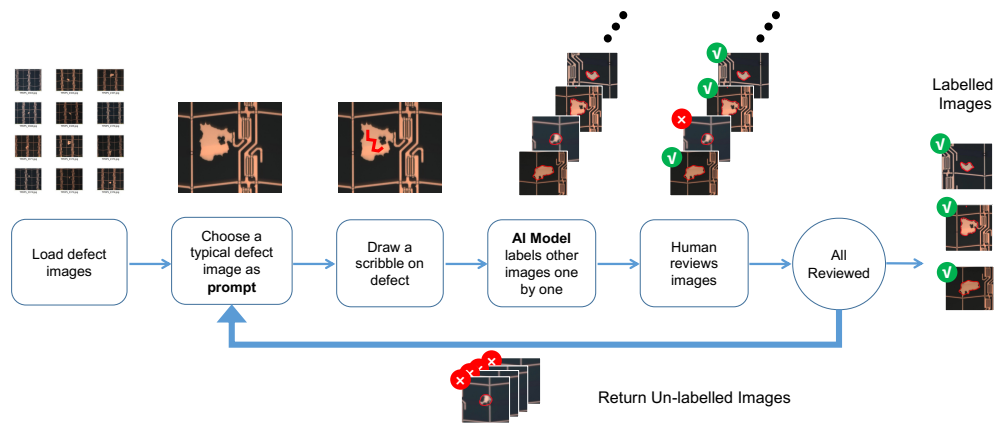

**Figure 1.** Auto-labelling flow using our proposed model

The remainder of this paper is organized as follows: Section 2 details our methodology and system architecture. Section 3 presents experimental results and analysis, followed by conclusions and future work in Section 5.

## 2. Methodology

***Auto-Labelling Workflow:*** Our proposed auto-labeling workflow, as shown in Figure 1, streamlines the annotation process through an iterative approach. Initially, a human annotator selects a representative defect image as a prompt and marks the defect region with a simple scribble. The AI model then uses this annotated prompt to automatically label similar defects across the remaining unlabeled images. A human expert reviews these AI-generated labels, approving accurate annotations and rejecting inaccurate ones. Successfully labeled images are added to the training dataset, while rejected images are returned to the unlabeled pool for reprocessing with different prompt images. This iterative feedback loop continues until all images are accurately labeled, creating an efficient annotation system that significantly reduces the manual effort required for dataset creation.

***Introduction to ICL Model:*** The SegGPT model employs a paired image processing methodology. The model processes reference-query image pairs, where each reference image is accompanied by its segmentation mask and paired with a corresponding query image and its segmentation mask. During training, the model leverages masked auto-encoding [8], where random regions across the paired segmentation masks are systematically occluded and replaced with placeholder tokens, establishing a self-supervised learning objective where the model learns to reconstruct these occluded regions. This training paradigm enables the model to develop robust contextual understanding and generalized segmentation capabilities. At inference time, the query image's segmentation mask is entirely replaced with placeholder tokens, prompting the model to generate the target segmentation mask based on the reference image's context, as illustrated in Figure 2. To adapt this architectural foundation for the specific challenges of industrial display defect segmentation, we introduce several domain-specific enhancements and architectural modifications, detailed in the following section.

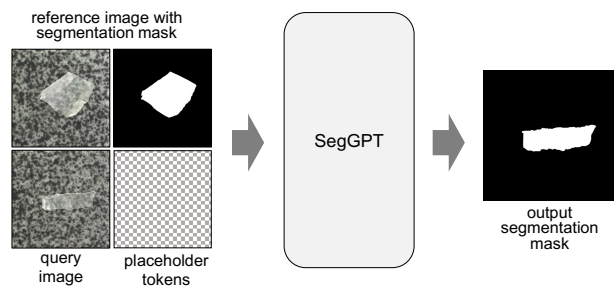

**Figure 2.** In-context learning based image segmentation via SegGPT

***Enhancing Defect Representation:*** To address the unique challenges of display panel defect detection, we implemented a sophisticated data representation strategy. Display panel defects typically occupy a small portion of the overall image, which can lead to inefficient learning and potential bias toward smaller defects during inference. Our solution introduces an adaptive cropping mechanism that captures defect areas with variable background margins, effectively creating a dynamic zoom effect. We further enhance the learning process by combining multiple zoomed-in defect images into a single composite frame, enabling the model to process multiple defects simultaneously. This approach, known as spatial ensembling [7], utilizes various grid configurations (1x1, 2x2, 3x3, and 4x4) while maintaining a balance with original small-defect images to ensure diverse learning scenarios. We found that configurations beyond 4x4 were impractical as they reduced the per-defect pixel count below the effective threshold for transformer patch processing.

***Multiclass Training for Robustness:*** While our auto-labeling implementation focuses on single-class segmentation during inference, we found that training the model with multiple defect classes simultaneously enhances its overall accuracy and robustness. Since our dataset already contains multiple defect classes, categorized based on their visual characteristics and root causes, we leverage SegGPT's inherent capability to handle multiple classes simultaneously through a color-coded approach. During training, each defect class is assigned a unique color representation across the RGB channels, with masks for different classes appearing together in training images. To prevent the model from developing color-specific biases, dynamic color assignment is used where mask colors are randomly selected for each training iteration. The color selection process employs systematic RGB color space partitioning, ensuring maximum color separation and uniqueness between different defect classes. This training approach ultimately improves the model's ability to distinguish and segment individual defect types.

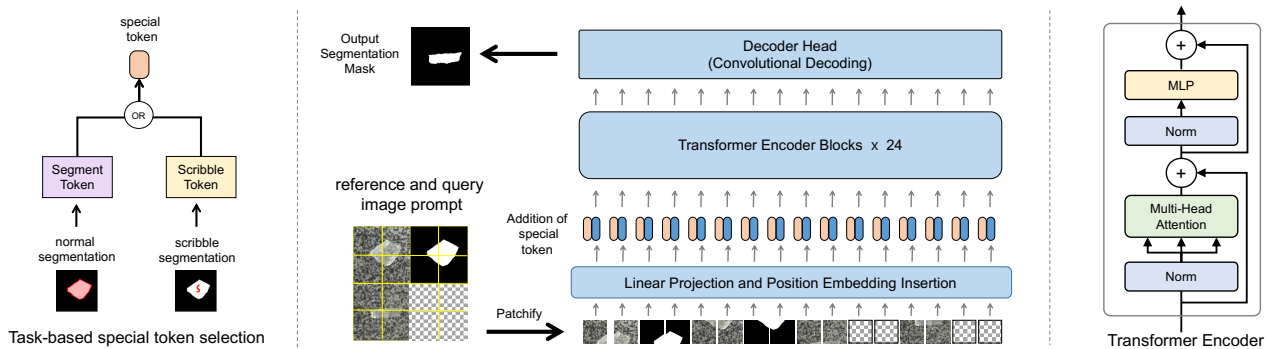

**Figure 3.** Modified architecture with special scribble token

***Selective Label Suppression:*** Through empirical observations, we identified a critical challenge: the model would attempt to segment any defect-like pattern it recognized from its training experience, regardless of the specific defect type indicated in the prompt. To enhance the model's prompt adherence and selective segmentation capabilities, we developed a label occlusion strategy. This approach involves deliberately suppressing labels of certain defect classes during training by randomly deactivating their annotations in the reference image while setting corresponding target masks in the query image to zero. This systematic label suppression teaches the model to discriminate between prompted and non-prompted defects, effectively training it to ignore defects that are not explicitly specified in the current prompt. The resulting enhancement in prompt faithfulness significantly improves the model's practical utility by ensuring it responds only to the defect types of immediate interest, rather than attempting to identify all potential defects simultaneously.

***Scribble-Based Annotation:*** Given the complex morphology and often ambiguous boundaries of display panel defects, creating precise segmentation masks during the annotation process can be challenging and time-consuming. Drawing inspiration from [9], we implemented a more practical approach that allows annotators to identify defects through simple scribbles or points, significantly streamlining the annotation workflow. To enable this functionality, we augmented the model architecture with a specialized learnable token that distinguishes between scribble-based and full-mask prompts. The model learns this scribble-to-mask correspondence through a dedicated training strategy where reference images are paired with identical query images, but the reference mask is replaced with a simplified scribble annotation. During these scribble-learning iterations, our custom scribble token supersedes the standard segmentation token used in mask-based prompting. This architectural modification, illustrated in Figure 3, enables the model to effectively translate simple user inputs into comprehensive segmentation masks, substantially reducing the annotation burden while maintaining segmentation accuracy.

## 3.  Experimental Results and Analysis

***Model Training:*** Our model training strategy employed a two-stage approach, building upon the pre-trained SegGPT model that was initially trained on diverse public domain datasets encompassing common objects and natural images. In the first stage, we fine-tuned the model on an extensive dataset of 200,000 LCD display panel circuit images with corresponding segmentation labels over 30 epochs. This stage incorporated our previously discussed augmentation and preprocessing techniques while excluding selective label suppression and scribble token functionality. The second stage involved further refinement using a carefully curated dataset of 10,000 high-quality images, where we consolidated visually similar defects into unified categories and filtered out extremely small defects that fell below our minimum pixel threshold. During this stage, we introduced both selective label suppression and scribble annotation capabilities, achieving optimal performance within just 10 epochs. The entire training pipeline was executed on a single node with four RTX 3090 consumer-grade GPUs with a combined video memory of 96 GB, demonstrating the accessibility of our approach using readily available hardware resources.

***Experiments and Evaluation:*** We conducted two sets of experiments to comprehensively evaluate our model's performance. The first set assessed the model's auto-labeling capabilities through four quantitative metrics: intersection-over-union (IoU), recall, Hausdorff distance (HD) [10], and coverage rate. IoU quantifies the overlap between predicted and ground truth segmentation masks, while recall represents the percentage of images where the model successfully detected defects with a positive IoU. The Hausdorff distance, which measures the shape similarity between predicted and ground truth masks, is defined as:

$$h(A,B) = \max_{a \in A} \min_{b \in B} |a - b|$$
$$H(A,B) = \max\{h(A,B), h(B,A)\} \quad (1)$$

where A and B are the sets of points in the predicted and ground truth masks respectively, and |a – b| is the Euclidean distance between points a and b. A lower HD value indicates better shape correspondence between the masks.

We also introduced a comprehensive quality metric called coverage rate, which represents the percentage of images meeting optimal segmentation criteria which requires IoU > 0.60 and HD < 10. As illustrated in Figure 4, our proposed model demonstrates substantial improvements over the pre-trained baseline across all product types with an average IoU increase of 0.22 and average recall increase of 14%. Figure 5 presents the coverage rate analysis, showing that the model, on average, successfully auto-labels approximately 60% of unseen data across three product categories, validating its generalization capabilities and the effectiveness of our approach.

To validate the practical utility of our system, we conducted a comparative analysis between auto-labeled and human-labeled data through a downstream task evaluation. As shown in Figure 6, we divided a set of human-labeled images into training and test groups. We then created two parallel evaluation streams: one using the original human labels for model training, and another using our auto-generated labels. Both resulting models were evaluated on the same test set. The results, presented in Table 1, demonstrate that models trained on auto-labeled data achieved comparable performance to those trained on human-labeled data, with nearly identical recall and accuracy metrics in the downstream segmentation task. These results prove that our auto-labeling approach can serve as an effective alternative to manual annotation in industrial inspection applications.

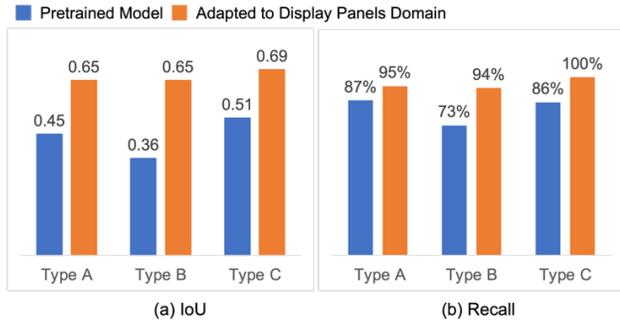

**Figure 4.** Performance of the proposed model on three types of LCD panels

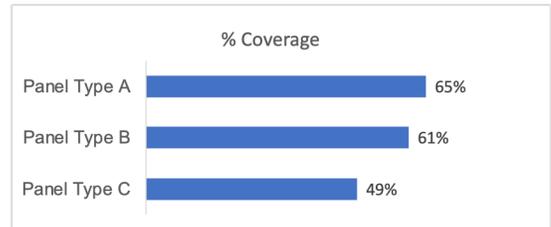

**Figure 5.** Coverage rate results on three types of LCD panels

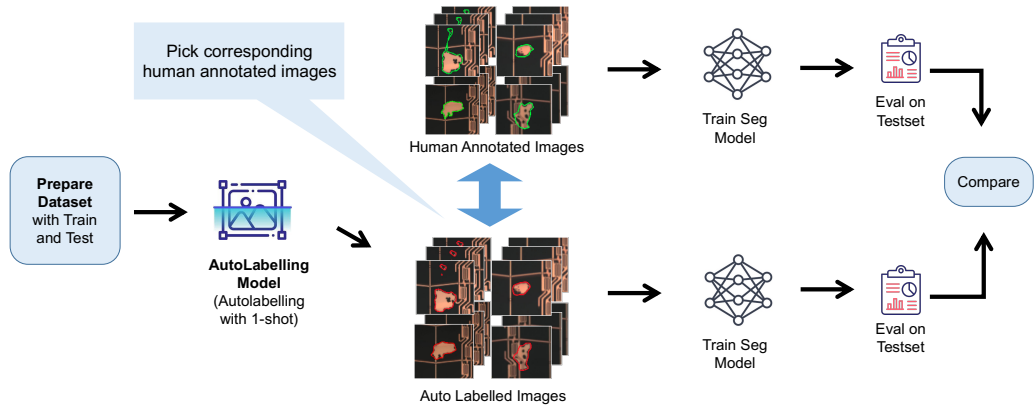

**Figure 6.** Experimental workflow to compare the quality of human-annotated and automatically labeled datasets.

**Table 1.** Performance comparison of model trained with human labelled data vs auto-labelled data

|  | **Human** | **Auto-labelled** |
|---|---|---|
| **IoU** | 0.85 | 0.84 |
| **Recall** | 98.83% | 99.16% |

## 4. Conclusion

In this paper, we present a deep learning-based approach to automated labeling for display panel manufacturing data using in-context learning through an enhanced SegGPT architecture. Our methodology addresses the critical bottleneck of manual annotation. Through strategic modifications—including a specialized training strategy optimized for display panel characteristics, an efficient scribble-based annotation mechanism, and selective label suppression for improved prompt adherence—we have developed a system that significantly streamlines the labeling process while maintaining high accuracy.

The experimental results demonstrate the effectiveness of our approach, achieving comparable performance to human annotations with approximately 60% auto-labeling coverage across diverse product categories. The integration of scribble-based input mechanisms has notably reduced the annotation burden while maintaining segmentation quality. Furthermore, our two-stage training strategy, combining extensive pre-training with focused fine-tuning on curated datasets, has proven effective in developing a robust and generalizable model.

This work represents a significant step toward practical, AI-assisted quality control in manufacturing environments and offers promising directions for future development. Future research could explore extending this methodology to other manufacturing domains and investigating ways to further reduce the need for human verification in the annotation process.